%% file: main.tex
\documentclass[sigconf]{acmart}

\usepackage{microtype}
\usepackage{graphicx}
\usepackage{subfigure}

\usepackage{hyperref}

\usepackage[font=small,skip=4pt]{caption}
\usepackage{amsmath}
 
\usepackage{amssymb}
\usepackage{mathtools}
\usepackage{amsthm}
\usepackage{algorithm, algpseudocode}

\usepackage[capitalize,noabbrev]{cleveref}
\usepackage{xspace}

\usepackage[nolist]{acronym}
\input{acronyms.tex}
\input{notation.tex}

\theoremstyle{plain}

\theoremstyle{definition}

\theoremstyle{remark}

\acmISBN{978-1-4503-XXXX-X/2018/06}

\usepackage[nolist]{acronym}
\title{Parameter Averaging in Link Prediction}
\author{Rupesh Sapkota}
\affiliation{%
\institution{Heinz Nixdorf Institute, Paderborn University}
\city{Paderborn}
\state{Nordrhein-Westfalen}
\country{Germany}}
\email{rupezzz@mail.uni-paderborn.de}

\author{Caglar Demir}
\affiliation{%
\institution{Heinz Nixdorf Institute, Paderborn University}
\city{Paderborn}
\state{Nordrhein-Westfalen}
\country{Germany}}
\email{caglar.demir@uni-paderborn.de}

\author{Arnab Sharma}
\affiliation{%
\institution{Heinz Nixdorf Institute, Paderborn University}
\city{Paderborn}
\state{Nordrhein-Westfalen}
\country{Germany}}
\email{arnab.sharma@uni-paderborn.de}

\author{Axel-Cyrille Ngonga Ngomo}
\affiliation{%
\institution{Heinz Nixdorf Institute, Paderborn University}
\city{Paderborn}
\state{Nordrhein-Westfalen}
\country{Germany}}
\email{axel.ngonga@upb.de}

 \copyrightyear{2025} 
\acmYear{2025} 
\setcopyright{cc}
 \setcctype{by}
 \acmConference[K-CAP '25]{Knowledge Capture Conference 2025}{December
 10--12, 2025}{Dayton, OH, USA}
 \acmBooktitle{Knowledge Capture Conference 2025 (K-CAP '25), December
10--12, 2025, Dayton, OH, USA}
 \acmDOI{10.1145/3731443.3771365}
 \acmISBN{979-8-4007-1867-0/2025/12}
\begin{document}

\begin{abstract}
Ensemble methods are widely employed to improve generalization in machine learning. This has also prompted the adoption of ensemble learning for the knowledge graph embedding (KGE) models in performing link prediction. Typical approaches to this end train multiple models as part of the ensemble, and the diverse predictions are then averaged. However, this approach has some significant drawbacks. For instance, the computational overhead of training multiple models increases latency and memory overhead. In contrast, model merging approaches offer a promising alternative that does not require training multiple models. In this work, we introduce model merging, specifically {\em weighted averaging}, in KGE models. Herein, a running average of model parameters from a training epoch onward is maintained and used for predictions. 
To address this, we additionally propose an approach that selectively updates the running average of the ensemble model parameters only when the generalization performance improves on a validation dataset. 
We evaluate these two different weighted averaging approaches on link prediction tasks, comparing the state-of-the-art benchmark ensemble approach.  Additionally, we evaluate the weighted averaging approach considering literal-augmented KGE models  and multi-hop query answering tasks as well.  The results demonstrate that the proposed weighted averaging approach consistently improves performance across diverse evaluation settings.

\end{abstract}



\maketitle

\section{Introduction}
%
Ensemble learning is one of the most effective techniques to improve the generalization of \ac{ML} algorithms across challenging tasks~\cite{dietterich2000ensemble,murphy2012machine,goodfellow2016deep}.
In its simplest form, an ensemble model is constructed from a set of $K$ different learners over the same training set~\cite{breiman1996bagging}. 
At test time, a new data point is classified by taking a (weighted) average of $K$
the predictions of the $K$ learners~\cite{sagi2018ensemble}.
Although prediction averaging often improves predictive accuracy, uncertainty estimation, and out-of-distribution robustness~\cite{garipov2018loss,liu2022deep},
it incurs computational overhead of training $K$ models, increased latency and memory requirements at test time~\cite{liu2022deep}.

In recent years, a number of works have considered employing the ensemble approach in knowledge graph embedding models~\cite{krompass2015ensemble,WanDPW20,xu2021multiple,rivas2022ensembles,GregucciN0S23,prabhakar2023ensemble}. One of the earliest work to this end is by Krompass et al.~\cite{krompass2015ensemble}, where multiple heterogeneous models are combined to get an ensemble of KGE models. The scores of these models are averaged to get the final prediction scores. A similar approach is also followed in~\cite{rivas2022ensembles,GregucciN0S23}. This leads to better accuracy, however, at the cost of training multiple models. There exist other approaches such as subgraph-based ensembling~\cite{WanDPW20,prabhakar2023ensemble}, low dimensional ensemble~\cite{xu2021multiple}, snpashot-based ensemble~\cite{huang2017snapshot,ShabanP24}. However, most of these approaches rely on explicit score-level aggregation or model duplication, which either require significant storage for storing full checkpoint ensembles and manual tuning using relation-aware weight assignment. This incurs additional inference overhead.

Recently, Izmaliov et al.~\cite{izmailov2018averaging} show that
\ac{SWA} technique often improves the generalization performance of a neural network by averaging $K$ \emph{snapshots} of parameters at the end of each epoch from a specific epoch onward.
Therefore, while a neural network is being trained (called an underlying running model), an ensemble model is constructed through averaging the trajectory of an underlying running model in a parameter space.
Therefore, at test time, the memory and running time requirements of using \ac{SWA} parameter ensemble are identical to the requirements of using a single neural network.
Although the idea of averaging parameters of linear models to accelerate \ac{SGD} on convex problems dates back to the works by Polyak et al.~\cite{polyak1992acceleration}, 
Izmailov et al.~\cite{izmailov2018averaging} show that an effective parameter ensemble of ML models can be built by solely maintaining a running average of parameters from a specific epoch onward.

In this work, we introduce the weighted averaging technique for the knowledge graph embedding models. Instead of training individual models, herein we obtain an ensemble of KGE models within a single training run. Therefore, unlike the existing ensembling approaches that require training and storing multiple independent models, our method maintains a running average of model weights throughout training, capturing diverse solutions in weight space. We found that selecting an unfittingly low number of training epochs leads  \ac{SWA} to suffer more from underfitting than an underlying running model.
In other words, a parameter ensemble model is heavily influenced by the early stages of training.

Therefore, striking the right balance in the choice of training epochs is crucial for harnessing the full potential of \ac{SWA} in enhancing the generalization. To address this, we propose \ac{ASWA} technique that extends SWA by building a parameter ensemble based on an adaptive schema governed by the generalization trajectory on the validation dataset.

\approach can be seen as a combination of \ac{SWA} with the early stopping technique,
as \ac{SWA} accepts all updates on a parameter ensemble, 
while the early stopping technique rejects any updates on a running model. 

To evaluate the effectiveness of both the weighted averaging techniques in KGE models, we perform
an extensive evaluation considering link prediction, literal score prediction, and multi-hop query answering tasks. The results of the evaluations suggest that 
weighted averaging improves the generalization performance over single and the existing state-of-the-art ensemble approaches. 
The main contributions of this paper are as follows:
\begin{enumerate}
    \item We introduce the weighted averaging approach in the knowledge graph embedding domain.
    \item We extend the stochastic weighted averaging approach by exploiting the early stopping technique. 
     \item We perform extensive evaluations considering different link prediction tasks to quantify the effectiveness of the weighted averaging approach. 
     \item We compare the effectiveness of the two weighted averaging approaches with diverse KGE ensemble models.
     \item We provide an open-source implementation of our approach.
\end{enumerate}

\section{Related Work}
\label{section:related_work}
Ensemble learning has been extensively studied in machine learning research~\cite{murphy2012machine}. At its core, an ensemble model is created by combining the predictions of a set of $K$ learners, often through simple averaging~\cite{breiman1996bagging}.
During inference, the final prediction is obtained by averaging the outputs of the $K$ learners, resulting in a more robust and accurate model.
For example, averaging the predictions of $K$ independently trained neural networks of the same architecture has been shown to significantly enhance test set performance~\cite{allen-zhu2023towards}. 

However, in the specific context of link prediction, ensemble-based methods have not been explored extensively~\cite{abs-2410-21163}. One of the earliest works by Krompass et al.~\cite{krompass2015ensemble} proposed {\em model-based ensembling} methods, wherein heterogeneous base models (e.g., TransE, RESCAL, neural architectures) are combined via averaging their prediction scores, achieving better accuracy but at the cost of training multiple full-sized models. Later works by Gregucci et al.~\cite{GregucciN0S23}, Rivas-Barragan et al.~\cite{rivas2022ensembles}, also follow similar approach. For instance, Building on this framework, Gregucci et al.~\cite{GregucciN0S23} proposed an attention‑based ensembling approach that learns to weight model contributions per query.  This yields an ensemble of models that outperforms individual embedding models and adapts flexibly to varied relation types.
In a biomedical context, Rivas-Barragan et al.~\cite{rivas2022ensembles} applied ensemble modeling to link‑prediction tasks in drug‑disease knowledge graphs. They trained ten different Knowledge Graph Embedding Models (KGEMs) and experimented with different ensemble strategies, such as  rank‑based and distribution‑based aggregation for combining predictions.

A different line of work on KGE ensemble focus on subgraph-based ensembling~\cite{prabhakar2023ensemble,WanDPW20}. This approach partitions the KG into subgraphs, trains separate embeddings on each, and aggregates results, however offering mixed evidence on computational efficiency.
Xu et al.~\cite{xu2021multiple} proposed {\em low-dimensional ensembling} where several instances of base models with smaller embedding sizes are trained independently and their scores averaged. This method achieves better performance under the same memory budget compared to a larger single model, however, it still requires separate training runs for each instance. 
Most recently, SnapE~\cite{ShabanP24} introduces a snapshot ensemble technique tailored for KG link prediction. By storing multiple model checkpoints during a single training loop with cyclical learning rates, SnapE obtains diverse base models at the cost of one training run. The authors therein showed the effectiveness and efficiency of this approach over the previous ensemble-based KGE models.
Typically, most of the ensemble techniques can lead to substantial improvements in generalization across various learning problems.
Yet, as $|\mathcal{E}| + |\mathcal{R}|$ grows, leveraging the prediction average technique becomes computationally prohibitive.
Attempts to alleviate the computational overhead of training multiple models have been extensively studied.
For instance, Xie et al.~\cite{xie2013horizontal} show that saving parameters of a neural network periodically during training and composing a final prediction via a voting schema improves the generalization performance.
Moreover, the dropout technique can also be seen as a form of ensemble learning~\cite{hinton2012improving,srivastava2014dropout}. 

More specifically, preventing the co-adaptation of parameters by stochastically forcing them to be zero can be seen as
a \emph{geometric averaging}~\cite{warde2013empirical,baldi2013understanding,goodfellow2016deep}. 
Similarly, Monte Carlo Dropout can be seen as a variant of the Dropout that is used to approximate model uncertainty without sacrificing either computational complexity or test accuracy \cite{gal2016dropout}. 
Garipov et al.~\cite{garipov2018loss} show that
optima of neural networks are connected by simple pathways having near constant training accuracy~\cite{draxler2018essentially}.
As an alternative to typical ensemble-based approaches, Izmailov et al.~\cite{izmailov2018averaging} proposed a model merging technique called Stochastic Weight Averaging (SWA) that builds a parameter ensemble model with \emph{almost} the same training and exact test time cost as a single model.
Note that all these works addressing issues with typical ensemble approaches consider neural network models. To the best of our knowledge, this work is the first to study weighted averaging of KGE model parameters. 

\section{Background}

We start by giving the formal definition of the KGE models, and then we introduce the weighted averaging approach. 

Let $\mathcal{E}$ be a set of entities and $\mathcal{R}$ a set of relations. A knowledge graph (KG) is defined as a set of triples:
\begin{equation}
\mathcal{G} := \{ (h, r, t) \in \mathcal{E} \times \mathcal{R} \times \mathcal{E} \},
\end{equation}
where each triple represents a relation $r$ between entities $h$ and $t$.
To enable learning, knowledge graph embedding (KGE) models map entities and relations into a continuous space $\mathbb{V}^d$ (e.g., $\mathbb{R}, \mathbb{C}, \mathbb{H}, \mathbb{O}$), yielding embeddings $\mathbf{E} \in \mathbb{V}^{|\mathcal{E}| \times d}$ and $\mathbf{R} \in \mathbb{V}^{|\mathcal{R}| \times d}$~\cite{balavzevic2019hypernetwork,YangYHGD14a,TrouillonWRGB16,TYL19}.
Typically, KGE models are used for link prediction~\cite{hogan2021knowledge}, i.e., their scoring function is $\scoreFunc_\Theta: \entities \times \relations \times \entities \mapsto \mathbb{R}$,
where $\Theta$ denotes parameters and often comprise $\mathbf{E}$, $\mathbf{R}$, and additional parameters (e.g., affine transformations, batch normalizations, convolutions). Given an assertion in the form of a triple $\triple{h}{r}{t} \in \entities \times \relations \times \entities$, a prediction $\hat{y}:=\scoreFunc_\Theta\triple{h}{r}{t}$ signals the likelihood of $\triple{h}{r}{t}$ being true~\cite{dettmers2018convolutional}.

\textbf{Stochastic Weight Averaging}
Izmailov et al.~\cite{izmailov2018averaging} proposed Stochastic Weight Averaging (SWA) that builds a parameter ensemble model having \emph{almost} the same training and test time cost of a single model. 
For the KGE models, we define the SWA ensemble parameter (i.e., embedding) update as follows.
\begin{equation}
  \Theta_{\text{SWA}} \leftarrow \frac{\Theta_{\text{SWA}} \cdot n_{\text{models}} + \Theta}{n_{\text{models}} + 1} \,    
\end{equation}
where $\Theta_{\text{SWA}}$ and $\Theta$ denote the embeddings of the ensemble model and the running model, respectively.
At each \ac{SWA} update, $n_{\text{models}}$ is incremented by 1. To be effective, finding a \emph{good} start epoch for SWA is important. Selecting an unfittingly low start point and total number of epochs can lead SWA to underfit. 
Similarly, starting \ac{SWA} only on the last few epochs does not lead to an improvement, as shown in \cite{izmailov2018averaging}. Therefore, it is important to perform weighted averaging only when it leads to performance improvement. Using this observation, we propose an {\em adaptive} version of the weighted averaging approach which we describe next.

\section{Adaptive Stochastic Weight Averaging}
\label{section:methodology}

While SWA averages embeddings uniformly from a fixed start epoch $j$, its effectiveness is highly sensitive to the choice of $j$: starting too early can result in underfitting, whereas averaging too late fails to improve generalization~\cite{izmailov2018averaging}. In contrast, our adaptive formulation \approach constructs the ensemble as a weighted sum of intermediate embeddings of the KGE model, which can be defined as follows.
\begin{equation}
\label{eq:aswa}
    \Theta_{\approach} = \sum_{i=1} ^N \pmb{\alpha}_i \odot \Theta_i,
\end{equation}
where $\Theta_i \in \mathbb{R}^d$ stands for a embedding vector of a running model at the $i-$th iteration.
$\odot$ denotes the scalar vector multiplication.
An ensemble coefficient $\pmb{\alpha}_i \in [0, 1]$ is a scalar value denoting the weight of $\Theta_i$ in $\Theta_\approach$.Here, $N$ denotes the number of epochs as does in \ac{SWA}~\cite{izmailov2018averaging}. Note that,  $\Theta_\approach$ corresponds to  $\Theta_{\text{SWA}}$ starting from the first iteration, if $\pmb{\alpha} = \mathbf{\frac{1}{N}}$, and $\Theta_\approach$ corresponds to  $\Theta_{\text{SWA}}$ starting from the $j$-th iteration if $\pmb{\alpha}_{0:j}= \mathbf{0}$ and $\pmb{\alpha}_{j+1:N}= \frac{\mathbf{1}}{\mathbf{N-j}}$.

We find that constructing an ensemble embedding model by averaging over multiple nonadjacent epoch intervals 
(e.g. two nonadjacent epoch intervals $i$ to $j$ and $k$ to m s.t. $m \ge k+1 \ge j+1 \ge i$) 
may be more advantageous than selecting a single start epoch. 
With these considerations, we argue that determining $\pmb{\alpha}$ according to an adaptive schema governed by the generalization trajectory on the validation dataset can be more advantageous than using a prefixed schema.
By this, we aim to ensure that $\Theta_\approach$ does not suffer
from underfitting more than the running embedding model $\Theta$ and overfitting more than $\Theta_{\text{SWA}}$.

Herein, the weights $\pmb{\alpha}_i$ for each individual embedding model  can be determined in a fashion akin to the early stopping technique~\cite{prechelt2002early}.
More specifically, the trajectory of the validation losses can be tracked at the end of each epoch.
By this, as the generalization performance degrades, the training can be stopped.
\begin{algorithm}[htb]
\caption{Adaptive Stochastic Weight Averaging (ASWA)}
\label{alg:ASWA}
\begin{algorithmic}[1] 
    \Require Initial model parameters $\Theta_0$, number of iterations $N$, training dataset $\mathcal{D}_{\text{train}}$, validation dataset $\mathcal{D}_{\text{val}}$
    \Ensure $\Theta_{\text{ASWA}}$
    \State $\Theta_{\text{ASWA}} \gets \Theta_0$ \Comment{Initialize Parameter Ensemble}
    \State $\pmb{\alpha} \gets \mathbf{0.0}$ \Comment{Initialize $N$ coefficients}
    \State $\text{val}_{\text{ASWA}} \gets -1$
    \Comment{Initialize validation score}
    \For{$i = 0$ to $N$}
        \State $\Theta_{i+1} \gets \Theta_i - \alpha \nabla \mathcal{L}(\Theta_i)$ 
        \State $\text{val}_{\Theta} \gets \text{Eval}(\mathcal{D}_{\text{val}}, \Theta_{i+1})$ 

        \If{$\text{val}_\Theta > \text{val}_{\text{ASWA}}$}
            \State $\Theta_{\text{ASWA}} \gets \Theta_{i+1}$ \Comment{Hard Update}
                \State $\pmb{\alpha} \gets \mathbf{0.0}$
                \State $\text{val}_{\text{ASWA}} \gets \text{val}_\Theta$

                \State \textbf{Continue}
        \EndIf

        \State $\hat{\Theta}_{\text{ASWA}} \gets \frac{\Theta_{\text{ASWA}} \odot (\sum_{j=i}^N \pmb{\alpha}_{j}) + \Theta_{i+1}}{(\sum_{j=i}^N \pmb{\alpha}_{j}) + 1}$  \Comment{Look-head}
        \State $\text{val}_{\hat{\Theta}_{\text{ASWA}}} \gets \text{Eval}(\mathcal{D}_{\text{val}}, \hat{\Theta}_{\text{ASWA}})$ 
        
        \If{$\text{val}_{\hat{\Theta}_{\text{ASWA}}} > \text{val}_{\text{ASWA}}$}
            \State $\Theta_{\text{ASWA}} \gets \hat{\Theta}_{\text{ASWA}}$ \Comment{Soft Update}
            \State $\pmb{\alpha}_{i+1} \gets 1.0$
            \State $\text{val}_{\text{ASWA}} \gets \text{val}_{\hat{\Theta}_{\text{ASWA}}}$
        \Else
            \State \Comment{Reject Update}
        \EndIf
    \EndFor
\end{algorithmic}
\end{algorithm}

In~\Cref{alg:ASWA}, we describe \approach with hard, soft ensemble updates and rejection according to the trajectories of the validation performances and an embedding ensemble model $\Theta_\approach$. 
By incorporating the validation performance of the running model, hard ensemble updates can be performed, i.e., if the validation performance of the running ensemble model is greater than the validation performances of the current ensemble model and the look-head (Lines 7-12). 
The hard ensemble update restarts the process of maintaining the running average of parameters, whereas the soft update updates the current parameter ensemble model based on the running model (Lines 15-18).
Note that \textbf{the validation performance of $\Theta_\approach$ cannot be less than the validation performance of $\Theta$}.
Hence, a possible underfitting depending on $N$ is mitigated with the expense of computing the validation performances.
Importantly, 
\textbf{the validation performance of $\Theta_\approach$ cannot be less than the validation performance of $\Theta_\text{SWA}$ due to the rejection criterion}, i.e., a parameter ensemble model is only updated if the validation performance is increased.

\paragraph{\textbf{Computational Complexity}} At test time, the time and memory requirements of \approach are identical to the requirements of conventional training as well as SWA.
Yet, during training, the time overhead of \approach is linear in the size of the validation dataset.
This stems from the fact that at each epoch the validation performances are computed.
Since \approach does not introduce any hyperparameter to be tuned, \approach can be more practical if the overall training and hyperparameter optimization phases are considered. 

\section{Experimental Setup}

\paragraph{Datasets}
In our experiments, we used the standard benchmark datasets  for link prediction such as UMLS, KINSHIP, Countries S1,   NELL-995 h50, NELL-995 h75, NELL-995 h100, FB15K-237, YAGO3-10 benchmark datasets. Apart from simple link predictions, we also consider multi-hop answering tasks where link predictors are being used.
Overviews of the datasets and different query types we consider are provided in~\Cref{table:datasets} and 
\Cref{tab:query-types}, respectively. Additionally, the literal extensions of FB15k-237 and YAGO15K contain 29,220 and 23,520 triples, and 116 and 7 attribute types, respectively.
\begin{table}[!htb]
    \caption{An overview of datasets in terms of the number of entities $\mathcal{E}$, number of relations $\mathcal{R}$, and the number of triples in each split of the knowledge graph datasets.}
    \centering
    \setlength{\tabcolsep}{1.8pt}
    \label{table:datasets}
    \begin{tabular}{lccccc}
    \toprule
    \textbf{Dataset} & \multicolumn{1}{c}{$|\entities|$}&  $|\relations|$ & $|\kg^{\text{Train}}|$ & $|\kg^{\text{Validation}}|$ &  $|\kg^{\text{Test}}|$\\
    \midrule
    Countries-S1    &271    &2  &1,111   &24 &24\\
    UMLS           &135      &46  &5,216      &652     &661\\
    KINSHIP        &104      &25  &8,544      &1,068   &1,074\\
    NELL-995-h100  &22,411   &43  &50,314     &3,763   &3,746\\
    NELL-995-h75   &28,085   &57  &59,135     &4,441   &4,389\\
    NELL-995-h50   &34,667   &86  &72,767     &5,440   &5,393\\
    WN18RR        & 40,943  &22  &86,835    &3,034   &3,134 \\
    YAGO15K        & 15,403 & 32  & 110,441 & 13,800 & 13,815 \\
    FB15K-237      &14,541   &237 &272,115    &17,535  &20,466 \\
     YAGO3-10       &123,182  &37  &1,079,040  &5,000   &5,000 \\
    \bottomrule
    \end{tabular}

\end{table}

\begin{table}[!htb]
\caption{Overview of different query types. 
Query types are taken from \cite{demir2023litcqd}.}
\label{tab:query-types}
    \centering
    \large
    \setlength{\tabcolsep}{1.0pt}
    
    \begin{tabular}{@{}ll@{}}
        \toprule
        \multicolumn{2}{@{}c@{}}{\bfseries Multi-hop Queries} \\
        \midrule

        2p            & $E_?\:.\:\exists E_1:r_1(e,E_1)\land r_2(E_1, E_?)$                                \\
        3p            & $E_?\:.\:\exists E_1E_2.r_1(e,E_1)\land r_2(E_1, E_2)\land r_3(E_2,E_?)$           \\
        2i            & $E_?\:.\:r_1(e_1,E_?)\land r_2(e_2,E_?)$                                           \\
        3i            & $E_?\:.\:r_1(e_1,E_?)\land r_2(e_2,E_?)\land r_3(e_3,E_?)$                         \\
        ip            & $E_?\:.\:\exists E_1.r_1(e_1,E_1)\land r_2(e_2,E_1)\land r_3(E_1,E_?)$             \\
        pi            & $E_?\:.\:\exists E_1.r_1(e_1,E_1)\land r_2(E_1,E_?)\land r_3(e_2,E_?)$             \\
        2u            & $E_?\:.\:r_1(e_1,E_?)\lor r_2(e_2,E_?)$                                            \\
        up            & $E_?\:.\:\exists E_1.[r_1(e_1,E_1)\lor r_2(e_2,E_1)]\land r_3(E_1,E_?)$\\
        \bottomrule
    \end{tabular}
\end{table}

\paragraph{Training and Optimization}
We followed standard experimental setups in our experiments.
For the link prediction and multi-hop query answering,
we followed the experimental setup used in~\cite{trouillon2016complex, ruffinelli2020you,arakelyan2021complex}. 
We trained DistMult, ComplEx, QMult, and Keci \ac{KGE} models with the following hyperparameter configuration: 
the number of epochs $N \in \{128, 256, 300\}$, Adam optimizer with $\eta=0.1$,
batch size $1024$, and an embedding vector size $d=128$.
Note that $d=128$ corresponds to 128 real-valued embedding vector size, hence 64 and 32 complex- and quaternion-valued embedding vector sizes, respectively.
We ensure that all models have the same number of parameters while exploring various $d$.
Throughout our experiments, we used the KvsAll training strategy.
We applied the beam search combinatorial search to apply pre-trained aforementioned \ac{KGE} models to answer multi-hop queries.
More specifically, we compute query scores for entities via the beam search combinatorial optimization procedure, we keep the top $10$ most promising variable-to-entity substitutions. Furthermore, we apply SWA from the starting epoch and use its best-found hyperparameters for \approach to ensure a fair comparison, as \approach does not rely on selecting a start point.

Similarly, for \textbf{SnapE}, we adopt the \textit{deferred CCA} scheduling method for cyclic learning rate adjustments, combined with a \textit{weighted averaging} of scores from an ensemble of models. The weights are assigned based on the inverse of the training loss at the time each snapshot model is captured. While the original work recommends deferring the learning rate scheduling until 80\% of the training is complete, we initiate the snapshot ensembling at approximately the halfway point of training. For the \textbf{literal-augmented models}, we maintain the same embedding dimension as used in other link prediction tasks ($d = 128$). We perform a hyperparameter search over learning rates $\in \{0.1, 0.01, 0.05, 0.001\}$ and number of epochs $\in \{256, 300, 500\}$, selecting the best configuration for all model variants within each dataset. Our goal is to evaluate model performance under different ensemble strategies rather than to identify the best-performing individual model.

\paragraph{Evaluation}
To evaluate the link prediction and multi-hop query answering performances, we used standard metrics filtered mean reciprocal rank (MRR) and Hit@1, Hit@3, and Hit@10~\cite{ruffinelli2020you}.
To evaluate the multi-hop query answering performances, we followed the complex query decomposition framework~\cite{arakelyan2021complex,demir2023litcqd}.
Therein, a complex multi-hop query is decomposed into subqueries, where the truth
value of each atom is computed by a pretrained KGE model/neural link predictor. 
Given a query, a prediction is obtained by
ranking candidates in descending order of their aggregated scores.
For each query type, we generate 500 queries to evaluate the performance. The implementation of the \approach, along with SWA and SnapE can be found here~\footnote{\url{https://github.com/dice-group/dice-embeddings}}.

\textbf{Link Prediction Results.} \Cref{table:wn18rr_fb15k237_yago310,table:lp_results_nell,table:lp_results_countries} report the link prediction results on WN18RR, FB15K-237, and YAGO3-10, NELL, Countries, UMLS, and KINSHIP datasets, respectively.
Overall, experimental results suggest that SWA and \approach consistently lead to better generalization performance in all metrics than conventional training and the state-of-the-art ensemble approach SnapE in the link prediction.
\input{tables/lp_wn_fb_yago}
\input{tables/lp_nell}
\input{tables/lP_countries_umls_kinship}
More specifically, \Cref{table:wn18rr_fb15k237_yago310} shows that SWA and \approach outperform the conventional training and SnapE on FB15k-237 and YAGO3-10 datasets.
On WN18RR, however, we see that SnapE performs the best with DistMult and QMult. This is because during deferred training, models show stable validation performance compared to other approaches while training loss decreases, indicating overfitting. In the ensemble, later snapshots get higher weights due to lower training loss; however, no better generalization, thereby slightly boosting performance through the weighting scheme.

Apart from these two, for all other models, our approach \approach shows MRR improvement over SWA in most cases. This is clearly visible on two large datasets, FB15k-237 and YAGO3-10. Specifically, on FB15K-237, \approach shows a notable advantage, outperforming SWA and SnapE for all models. On YAGO3-10, we see a similar trend—\approach significantly improves performance on most metrics, especially with ComplEx and Keci models.
\Cref{table:lp_results_nell} shows results on the NELL h100, h75, and h50 datasets. Again, the weighted averaging approaches outperform the best KGE ensemble approach. Moreover, \approach consistently leads to further performance gains across all models.
Finally, \Cref{table:lp_results_countries} shows a similar trend for smaller datasets. Apart from the Countries-S1 dataset with the ComplEx model, all other models show performance improvement with our weighted averaging approach.

\textbf{Literal Augmented Models.} To evaluate the effectiveness of weighted averaging approaches across various types of link prediction tasks, we also assess their performance using literal-augmented KGE models. Among existing methods, we adopt LiteralE~\cite{kristiadi2019incorporating} to incorporate literal information into KGE models and evaluate them independently, as well as with SWA and ASWA. For a comprehensive evaluation, we include geometric models (ComplEx~\cite{trouillon2016complex}, DistMult~\cite{yang2015embedding}), a convolution-based model (ConvE~\cite{dettmers2018convolutional}), and a tensor factorization model (TuckER~\cite{balazevic2019tucker}) and report the results in Table~\ref{table:lp_results_literals}. The results show that the weighted averaging approaches outperform the existing models. Furthermore, we can see that there is a clear advantage of using our weighted averaging approach \approach therein since it improves the results of the SWA for all the models. 
\input{tables/literal_LP}

\textbf{Multi-hop Query Answering Results}
Table~\ref{table:query_answering_fb_results} presents Mean Reciprocal Rank (MRR) results on FB15k-237 for a diverse set of multi-hop logical queries.
Overall, \approach demonstrates consistent improvements across models and query types, particularly improving the performance in queries involving intersection (3i) and union (2u, up). For instance, QMult+\approach achieves the highest MRR on 3i (0.183) and 2u (0.072), while Keci+\approach achieves strong gains across all complex queries.
While SWA generally boosts performance over the base models—especially on 3i queries—it exhibits inconsistent or even degraded performance on simpler path-based queries like 2p and 3p. This degradation is especially notable in DistMult and ComplEx, where MRR drops significantly (e.g., 2p drops from 0.007 to 0.002 in ComplEx+SWA).
These observations highlight the importance of adaptive or selective model averaging strategies—as implemented in \approach.

\begin{table}[!htb]
\centering
\small
\caption{Multi-hop query MRR results on FB15k-237.}
\label{table:query_answering_fb_results}
\setlength{\tabcolsep}{2.8pt}
\begin{tabular}{lcccccccccc}
        \toprule
& \textbf{2p}&\textbf{3p}&\textbf{3i}       & \textbf{ip}       & \textbf{pi}       & \textbf{ 2u}       & \textbf{up}       \\
        \midrule
DistMult           &\textbf{0.007} &\textbf{0.007}&0.044 & 0.003 &0.097&0.020&0.006 \\
+SWA       &0.003 &0.002 &0.106&0.004&\textbf{0.098}&0.031&\textbf{0.007} \\
+\approach &0.002&0.003&\textbf{0.175}&\textbf{0.005}&0.092&\textbf{0.055}&0.002\\
\midrule
ComplEx            &\textbf{0.009}&0.001&0.036&0.003&0.092&0.014&\textbf{0.006}\\
+SWA        &0.002&0.002&0.136&\textbf{0.005}&\textbf{0.105}&0.030&0.005\\
+\approach  &0.002&\textbf{0.003}&\textbf{0.155}&\textbf{0.005}&0.100&\textbf{0.049}&0.001\\
\midrule
QMult              &0.002&0.003&0.034&0.000&0.099&0.014&0.006\\
+SWA          &0.003&0.002&0.089&0.001&0.092&0.027&\textbf{0.007}\\
+\approach    &\textbf{0.004}&\textbf{0.005}&\textbf{0.183}&\textbf{0.002}&\textbf{0.117}&\textbf{0.072}&0.003\\
\midrule
Keci              &\textbf{0.009}&0.005&0.052&0.002&0.085&0.027&0.013\\
+SWA          &0.007         &0.005&0.101&0.002&0.097&0.059&\textbf{0.014}\\
+\approach    &0.004         &0.005&\textbf{0.165}&\textbf{0.007}&\textbf{0.106}&\textbf{0.064}&0.004\\
\bottomrule
\end{tabular}
\end{table}

\section{Discussion}

Our evaluations indicate that weight averaging approaches perform better than conventional training (e.g. finding model parameters with
Adam optimizer); moreover, \approach  consistently generalizes better than SWA  across models and datasets.
For instance, on YAGO3-10 (see Table~\ref{table:wn18rr_fb15k237_yago310}), SWA finds model parameters leading to higher link prediction performance on the training data in all metrics. 
Given that \ac{KGE} models have the same number of parameters (entities and relations are represented with $d$ number of real numbers), this superior performance of SWA over the conventional training can be explained as the mitigation of the noisy parameter updates around minima.
More specifically, maintaining a running unweighted average of parameters  becomes particularly useful around a minima by means of reducing the noise in the gradients of loss w.r.t. parameters that is caused by the mini-batch training.
\begin{figure}[!t]
\caption{Training and validation MRR scores of SWA and ASWA across epochs for different models on UMLS and KINSHIP datasets.}
    \label{fig:swa_vs_aswa_plot}
    \centering
    \includegraphics[width=1.0\linewidth]{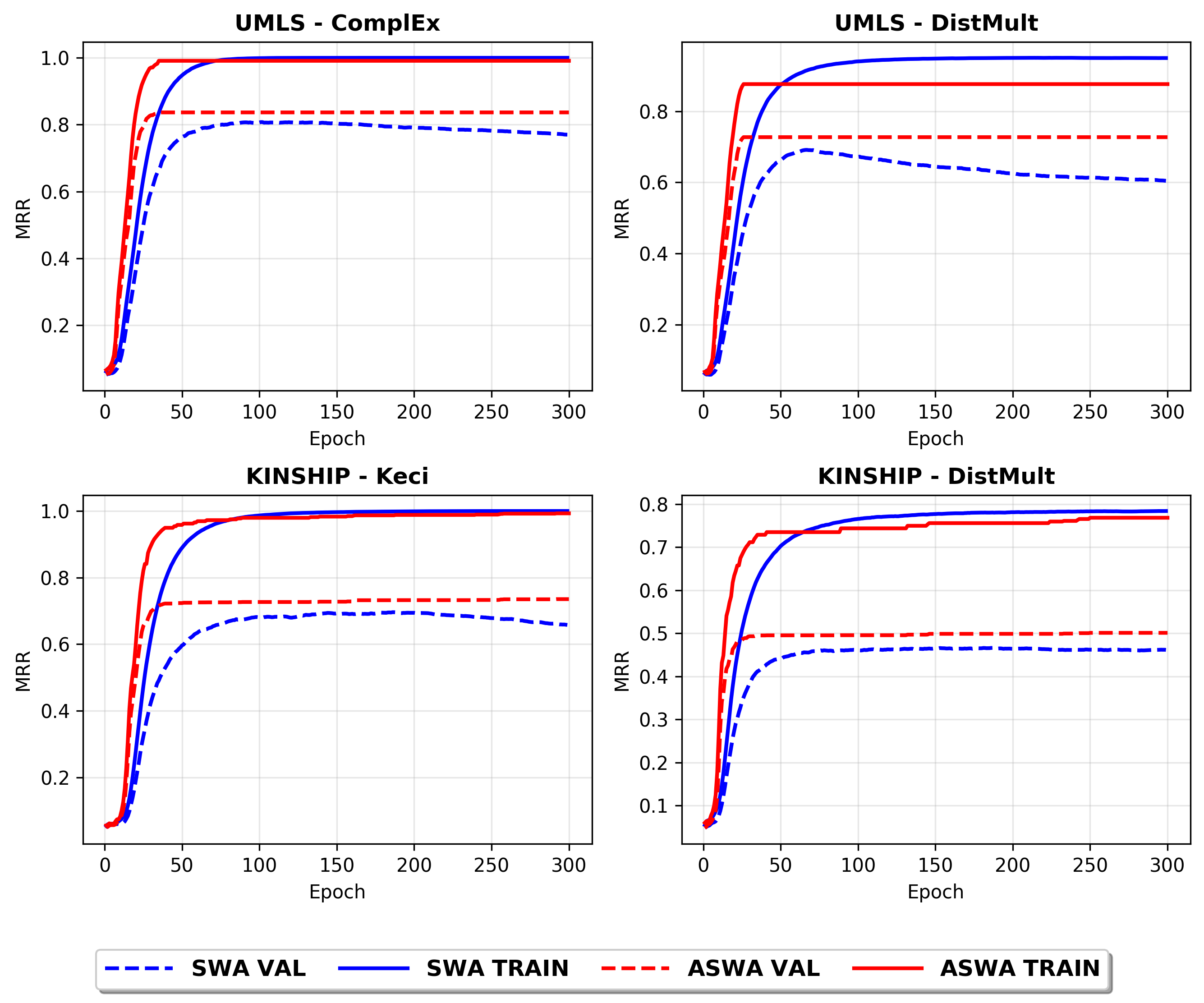}
    
\end{figure}
\approach, however, renders itself as an effective combination of SWA with early stopping techniques, where the former accepts all parameter updates on a parameter ensemble model based on a running model and the former rejects parameter updates on a running model in the presence of overfitting.
Since the \approach does not update a parameter ensemble model if a running model begins to overfit, it acts as a regularization on a parameter ensemble.
This regularization impact leads to a better generalization in all metrics.

Broadly, our experimental results corroborate the findings of \cite{izmailov2018averaging}: constructing a parameter ensemble by maintaining a running average of parameters at each epoch improves generalization across a wide range of datasets and models.
Link prediction and multi-hop query answering results show that SWA and \approach find better solutions than conventional training based on ADAM and SGD. Our results also show that updating the parameter ensemble uniformly at each epoch leads to suboptimal results as the model begins to overfit. This is clearly seen in Figure~\ref{fig:swa_vs_aswa_plot}, which shows the validation MRR comparing SWA and \approach as training progresses.
The results suggest that \approach effectively mitigates overfitting by rejecting parameter updates once the model begins to overfit. Thus, it serves as an effective combination of SWA and early stopping—SWA accepts all parameter updates, while early stopping rejects updates when overfitting occurs.
Importantly, \approach does not require a starting epoch to begin constructing the parameter ensemble. Instead, it performs a hard update on the ensemble model whenever the running model outperforms it on the validation dataset.

\section{Conclusion}

In this work, we investigated techniques to construct a high performing ensemble model while alleviating the overhead of training multiple models, and retaining efficient memory and inference requirements at test time.
To this end, we introduce the weighted averaging technique, SWA and furthermore 
propose an adaptive stochastic weight averaging technique that effectively combines the SWA technique with early stopping.
Essentially. \approach extends SWA by building a parameter ensemble according to an adaptive schema governed by the generalization trajectory on the validation dataset.
Our extensive experiments across benchmark datasets—spanning link prediction, literal-augmented models, and multi-hop query answering—demonstrate that weighted averaging techniques consistently improve the generalization performance of strong baseline models. Moreover, we observe that \approach is more effective than SWA in mitigating overfitting.
As future work, we aim to explore task-aware scheduling strategies for averaging that dynamically adjust based on query complexity or domain characteristics. Additionally, integrating uncertainty estimation into the averaging process could further enhance robustness and calibration considering noisy or adversarial settings.

\begin{acks}
    This work is supported by the Ministry of Culture and Science of North Rhine-Westphalia (MKW NRW) within the project SAIL under the grant no NW21-059D, the project WHALE (LFN 1-04) funded under the Lamarr Fellow Network Programme by the MKW NRW, the European Union’s Horizon Europe research and innovation programme under grant agreement No 101070305, and by the German Federal Ministry of Research, Technology and Space (BMFTR) within the project KI-OWL under the grant no 01IS24057B.
\end{acks}

\bibliographystyle{ACM-Reference-Format}
\bibliography{ref}
\end{document}

%% file: acronyms.tex
\begin{acronym}[UML]
	\acro{AI}{Artificial Intelligence}
 	\acro{ASWA}{Adaptive Stochastic Weight Averaging}

	\acro{CL}{Concept Learning}
	\acro{CEL}{Class Expression Learning}
	\acro{CWA}{Close World Assumption}
	\acro{CSV}{Comma-Separated Values}
	\acro{CNN}{Convolutional Neural Network}
	\acro{DL}{Description Logic}
	\acro{DL}{Deep Learning}
	\acro{DNN}{Deep Neural Network}
    \acro{FOL}{First-Order Logic}
  	\acro{GD}{Gradient Descent}
    \acro{KB}{Knowledge Base}
    \acro{KG}{Knowledge Graph}
    \acro{KGE}{Knowledge Graph Embedding}
	\acro{SW}{Semantic Web}
	\acro{OWA}{Open World Assumption}
	\acro{SGD}{Stochastic Gradient Descent}
 	\acro{SWA}{Stochastic Weight Averaging}
  
	\acro{MAP}{Maximum a Posteriori Probability Estimation}
	\acro{MRR}{Mean Reciprocal Rank}
	\acro{ML}{Machine Learning}
    \acro{MLE}{Maximum Likelihood Estimation}
    \acro{MDP}{Maximum Likelihood Estimation}
	\acro{MTL}{Multi-task Learning}
	\acro{MDP}{Markov Decision Process}
	\acro{NN}{Neural Network}
    \acro{KG}{Knowledge Graph}
    \acro{KGE}{Knowledge Graph Embedding}
    \acro{FGE}{Fast Geometric Ensembling}
\end{acronym}  

%% file: notation.tex
\newcommand{\approach}{\textsc{ASWA}\xspace}

\newcommand{\triple}[3]{(\texttt{#1}, \texttt{#2}, \texttt{#3})}

\newcommand{\kg}{\ensuremath{\mathcal{G}}\xspace}
\newcommand{\entities}{\ensuremath{\mathcal{E}}\xspace}
\newcommand{\relations}{\ensuremath{\mathcal{R}}\xspace}

\newcommand{\scoreFunc}{\phi}

%% file: tables/lp_wn_fb_yago.tex
\begin{table*}[htb]

\caption{ 
The table presents the link prediction performance of DistMult, ComplEx, QMult, and Keci, alongside their SWA and \approach variants, across 3 benchmark datasets: WN18RR, FB15K-237, and YAGO3-10. 
Each model is evaluated using Mean Reciprocal Rank (MRR) and Hits@k metrics (for $k = 1, 3, 10$). 
Bold values represent the best scores for each metric, indicating the top-performing model for that dataset.}

\label{table:wn18rr_fb15k237_yago310}
\centering
 \resizebox{0.93\textwidth}{!}{\begin{tabular}{l  cccc cccc cccc}
\toprule
  &\multicolumn{4}{c}{\textbf{WN18RR}} & \multicolumn{4}{c}{\textbf{FB15K-237}} & \multicolumn{4}{c}{\textbf{YAGO3-10}}\\
  \cmidrule(l){2-5} \cmidrule(l){6-9} \cmidrule(l){10-13}
                   & MRR  & @1     &@3     & @10 & MRR  & @1     &@3     & @10       & MRR  & @1 & @3 & @10\\
\toprule
ComplEx & 0.279 & 0.249 & 0.298 & 0.332 & 0.095 & 0.060 & 0.099 & 0.162 & 0.133 & 0.087 & 0.149 & 0.218 \\
+SWA & 0.288 & 0.262 & 0.303 & 0.332 & 0.153 & 0.109 & 0.162 & 0.233 & 0.459 & 0.395 & 0.502 & 0.571 \\
+SnapE & 0.284 & 0.256 & 0.302 & 0.334 & 0.155 & 0.102 & 0.169 & 0.256 & 0.403 & 0.332 & 0.449 & 0.531 \\
+ASWA & \textbf{0.351} & \textbf{0.340} & \textbf{0.358} & \textbf{0.370} & \textbf{0.192} & \textbf{0.140} & \textbf{0.207} & \textbf{0.289} & \textbf{0.470} & \textbf{0.403} & \textbf{0.515} & \textbf{0.585} \\
\midrule
DistMult & 0.281 & 0.265 & 0.290 & 0.309 & 0.092 & 0.062 & 0.095 & 0.149 & 0.131 & 0.097 & 0.146 & 0.192 \\
+SWA & 0.360 & 0.352 & 0.362 & 0.374 & 0.150 & 0.104 & 0.162 & 0.235 & \textbf{0.466} & \textbf{0.399} & \textbf{0.513} & \textbf{0.582} \\
+SnapE & \textbf{0.363} & \textbf{0.353} & \textbf{0.365} & \textbf{0.381} & 0.148 & 0.095 & 0.159 & 0.252 & 0.329 & 0.247 & 0.376 & 0.488 \\
+ASWA & 0.354 & 0.349 & 0.356 & 0.363 & \textbf{0.207} & \textbf{0.152} & \textbf{0.222} & \textbf{0.313} & 0.369 & 0.293 & 0.408 & 0.514 \\
\midrule
QMult & 0.085 & 0.065 & 0.090 & 0.121 & 0.111 & 0.080 & 0.117 & 0.161 & 0.268 & 0.211 & 0.288 & 0.378 \\
+SWA & 0.051 & 0.033 & 0.054 & 0.084 & 0.174 & 0.123 & 0.188 & 0.269 & 0.275 & 0.214 & 0.302 & 0.389 \\
+SnapE & \textbf{0.104} & \textbf{0.076} & \textbf{0.115} & \textbf{0.159} & 0.151 & 0.108 & 0.155 & 0.221 & 0.304 & 0.245 & 0.329 & 0.411 \\
+ASWA & 0.088 & 0.068 & 0.094 & 0.123 & \textbf{0.241} & \textbf{0.176} & \textbf{0.262} & \textbf{0.366} & \textbf{0.358} & \textbf{0.284} & \textbf{0.395} & \textbf{0.497} \\
\midrule
Keci & 0.305 & 0.275 & 0.324 & 0.359 & 0.128 & 0.079 & 0.137 & 0.224 & 0.213 & 0.148 & 0.240 & 0.342 \\
+SWA & 0.326 & 0.315 & 0.331 & 0.346 & 0.156 & 0.103 & 0.166 & 0.259 & 0.417 & 0.336 & 0.464 & 0.567 \\
+SnapE & 0.333 & 0.317 & 0.343 & 0.360 & 0.224 & 0.149 & 0.246 & \textbf{0.375} & 0.397 & 0.317 & 0.444 & 0.546 \\
+ASWA & \textbf{0.358} & \textbf{0.352} & \textbf{0.359} & \textbf{0.368} & \textbf{0.243} & \textbf{0.177} & \textbf{0.262} & 0.374 & \textbf{0.459} & \textbf{0.375} & \textbf{0.511} & \textbf{0.607} \\
\bottomrule
\end{tabular}}
\end{table*}

%% file: tables/lp_nell.tex
\begin{table*}[ht]
\caption{
The table reports the performance of different models, including DistMult, ComplEx, QMult, and Keci, along with their SWA and \approach variants, on the h100, h75, and h50 of NELL-995 benchmark dataset. Each model is evaluated using Mean Reciprocal Rank (MRR) and Hits@k metrics (for $k = 1, 3, 10$). 
Bold values represent the best scores for each metric, indicating the top-performing model for that dataset.}

\label{table:lp_results_nell}
\centering
 \resizebox{0.95\textwidth}{!}{\begin{tabular}{l  cccc cccc cccc}
\toprule
  &\multicolumn{4}{c}{\textbf{NELL-995-h100}} & \multicolumn{4}{c}{\textbf{NELL-995-h75}} & \multicolumn{4}{c}{\textbf{NELL-995-h50}}\\
  \cmidrule(l){2-5} \cmidrule(l){6-9} \cmidrule(l){10-13}
                   & MRR  & @1     &@3     & @10 & MRR  & @1     &@3     & @10       & MRR  & @1 & @3 & @10\\

\toprule
ComplEx & 0.192 & 0.135 & 0.212 & 0.308 & 0.192 & 0.137 & 0.210 & 0.301 & 0.195 & 0.142 & 0.214 & 0.300 \\
+SWA & 0.212 & \textbf{0.154} & \textbf{0.236} & 0.326 & 0.214 & 0.156 & 0.237 & 0.325 & 0.212 & 0.158 & 0.234 & 0.317 \\
+SnapE & 0.197 & 0.139 & 0.218 & 0.314 & 0.198 & 0.144 & 0.217 & 0.308 & 0.199 & 0.145 & 0.219 & 0.307 \\
+ASWA & \textbf{0.213} & 0.151 & 0.233 & \textbf{0.334} & \textbf{0.218} & \textbf{0.158} & \textbf{0.242} & \textbf{0.335} & \textbf{0.232} & \textbf{0.172} & \textbf{0.256} & \textbf{0.347} \\
\midrule
DistMult & 0.132 & 0.094 & 0.146 & 0.203 & 0.110 & 0.076 & 0.119 & 0.177 & 0.121 & 0.084 & 0.137 & 0.192 \\
+SWA & 0.175 & 0.129 & 0.193 & 0.262 & 0.162 & 0.122 & 0.179 & 0.237 & 0.158 & 0.115 & 0.177 & 0.242 \\
+SnapE & 0.166 & 0.115 & 0.187 & 0.262 & 0.150 & 0.108 & 0.163 & 0.234 & 0.154 & 0.106 & 0.176 & 0.243 \\
+ASWA & \textbf{0.238} & \textbf{0.172} & \textbf{0.265} & \textbf{0.376} & \textbf{0.229} & \textbf{0.166} & \textbf{0.257} & \textbf{0.352} & \textbf{0.239} & \textbf{0.177} & \textbf{0.268} & \textbf{0.360} \\
\midrule
QMult & 0.149 & 0.099 & 0.163 & 0.247 & 0.160 & 0.110 & 0.174 & 0.260 & 0.153 & 0.104 & 0.168 & 0.252 \\
+SWA & 0.168 & 0.114 & 0.186 & 0.271 & 0.176 & 0.124 & 0.195 & 0.279 & 0.162 & 0.114 & 0.176 & 0.260 \\
+SnapE & 0.108 & 0.072 & 0.117 & 0.179 & 0.112 & 0.079 & 0.119 & 0.173 & 0.118 & 0.081 & 0.125 & 0.196 \\
+ASWA & \textbf{0.176} & \textbf{0.118} & \textbf{0.196} & \textbf{0.287} & \textbf{0.183} & \textbf{0.126} & \textbf{0.205} & \textbf{0.292} & \textbf{0.174} & \textbf{0.120} & \textbf{0.191} & \textbf{0.278} \\
\midrule
Keci & 0.155 & 0.109 & 0.167 & 0.250 & 0.150 & 0.107 & 0.159 & 0.237 & 0.173 & 0.121 & 0.188 & 0.278 \\
+SWA & 0.204 & \textbf{0.152} & 0.223 & 0.310 & 0.195 & 0.144 & 0.212 & 0.298 & 0.217 & 0.161 & 0.238 & 0.332 \\
+SnapE & 0.164 & 0.115 & 0.180 & 0.264 & 0.158 & 0.114 & 0.168 & 0.251 & 0.180 & 0.129 & 0.194 & 0.285 \\
+ASWA & \textbf{0.212} & 0.150 & \textbf{0.235} & \textbf{0.341} & \textbf{0.230} & \textbf{0.165} & \textbf{0.256} & \textbf{0.357} & \textbf{0.247} & \textbf{0.184} & \textbf{0.273} & \textbf{0.377} \\

\bottomrule
\end{tabular}}
\end{table*}

%% file: tables/lP_countries_umls_kinship.tex
\begin{table*}

\caption{
The table presents the link prediction performance of various models, including DistMult, ComplEx, QMult, and Keci, along with their SWA and \approach variants, on the Countries-S1, UMLS and KINSHIP datsets.
The evaluation is conducted using Mean Reciprocal Rank (MRR) and Hits@k (for $k = 1, 3, 10$). 
Bold values represent the best scores for each metric, indicating the top-performing model for that dataset.}

\label{table:lp_results_countries}
\centering
\small
\resizebox{0.9\textwidth}{!}{\begin{tabular}{l  cccc cccc cccc}
\toprule
  &\multicolumn{4}{c}{\textbf{S1}} & \multicolumn{4}{c}{\textbf{UMLS}} & \multicolumn{4}{c}{\textbf{KINSHIP}}\\
  \cmidrule(l){2-5} \cmidrule(l){6-9} \cmidrule(l){10-13}
                   & MRR  & @1     &@3     & @10 & MRR  & @1     &@3     & @10       & MRR  & @1 & @3 & @10\\
\toprule
ComplEx & \textbf{0.360} & \textbf{0.208} & 0.396 & \textbf{0.771} & 0.650 & 0.514 & 0.741 & 0.920 & 0.568 & 0.398 & 0.683 & 0.908 \\
+SWA & 0.284 & 0.146 & 0.292 & 0.667 & 0.782 & 0.693 & 0.843 & 0.948 & 0.667 & 0.524 & 0.772 & 0.926 \\
+SnapE & 0.350 & 0.188 & \textbf{0.417} & 0.729 & 0.664 & 0.530 & 0.753 & 0.923 & 0.566 & 0.392 & 0.684 & 0.909 \\
+ASWA & 0.339 & \textbf{0.208} & 0.354 & 0.646 & \textbf{0.837} & \textbf{0.757} & \textbf{0.906} & \textbf{0.974} & \textbf{0.744} & \textbf{0.616} & \textbf{0.849} & \textbf{0.962} \\
\midrule
DistMult & 0.379 & 0.229 & 0.479 & 0.667 & 0.434 & 0.300 & 0.477 & 0.737 & 0.388 & 0.239 & 0.422 & 0.755 \\
+SWA & 0.362 & 0.188 & 0.479 & 0.667 & 0.600 & 0.489 & 0.646 & 0.833 & 0.474 & 0.324 & 0.520 & 0.844 \\
+SnapE & 0.405 & 0.250 & \textbf{0.500} & \textbf{0.708} & 0.452 & 0.303 & 0.505 & 0.769 & 0.429 & 0.273 & 0.482 & 0.816 \\
+ASWA & \textbf{0.416} & \textbf{0.312} & 0.438 & 0.688 & \textbf{0.727} & \textbf{0.626} & \textbf{0.784} & \textbf{0.927} & \textbf{0.501} & \textbf{0.345} & \textbf{0.563} & \textbf{0.865} \\
\midrule
QMult & 0.365 & 0.208 & 0.479 & 0.625 & 0.653 & 0.528 & 0.728 & 0.899 & 0.532 & 0.365 & 0.632 & 0.876 \\
+SWA & \textbf{0.457} & \textbf{0.333} & \textbf{0.562} & 0.646 & 0.767 & 0.671 & 0.834 & 0.937 & 0.610 & 0.459 & 0.704 & 0.907 \\
+SnapE & 0.346 & 0.167 & 0.479 & 0.667 & 0.664 & 0.542 & 0.735 & 0.904 & 0.533 & 0.368 & 0.628 & 0.880 \\
+ASWA & 0.368 & 0.188 & 0.500 & \textbf{0.729} & \textbf{0.824} & \textbf{0.734} & \textbf{0.897} & \textbf{0.970} & \textbf{0.688} & \textbf{0.553} & \textbf{0.786} & \textbf{0.937} \\
\midrule
Keci & 0.268 & 0.125 & 0.292 & 0.604 & 0.551 & 0.412 & 0.613 & 0.836 & 0.495 & 0.313 & 0.607 & 0.872 \\
+SWA & 0.308 & \textbf{0.146} & 0.375 & 0.625 & 0.693 & 0.589 & 0.753 & 0.898 & 0.686 & 0.534 & 0.800 & 0.954 \\
+SnapE & 0.272 & 0.125 & 0.292 & 0.604 & 0.553 & 0.413 & 0.621 & 0.840 & 0.505 & 0.327 & 0.606 & 0.874 \\
+ASWA & \textbf{0.318} & \textbf{0.146} & \textbf{0.396} & \textbf{0.667} & \textbf{0.830} & \textbf{0.750} & \textbf{0.894} & \textbf{0.958} & \textbf{0.713} & \textbf{0.580} & \textbf{0.811} & \textbf{0.961} \\
\bottomrule
\end{tabular}}
\end{table*}

%% file: tables/literal_LP.tex
\begin{table}

\caption{
The table presents the link prediction performance of various models augmented with literal values, including DistMult, ComplEx, ConvE, along with their SWA and \approach variants. 
Bold values represent the best scores for each metric, indicating the top-performing model for that dataset.}
\label{table:lp_results_literals}
\centering
\small
\setlength{\tabcolsep}{2pt}
\begin{tabular}{l  cccc cccc}
\toprule
  &\multicolumn{4}{c}{\textbf{FB15k-237 + Literal}} & \multicolumn{4}{c}{\textbf{YAGO15K + Literal}} \\
  \cmidrule(l){2-5} \cmidrule(l){6-9} 
                   & MRR  & @1     &@3     & @10 & MRR  & @1     &@3     & @10    \\
\toprule

ComplEx & 0.261 & 0.182 & 0.284 & 0.418 & 0.309 & 0.235 & 0.339 & 0.452 \\
+SWA & 0.270 & 0.189 & 0.293 & 0.432 & 0.263 & 0.177 & 0.299 & 0.429 \\
+ASWA & \textbf{0.291} & \textbf{0.209} & \textbf{0.317} & \textbf{0.454} & \textbf{0.412} & \textbf{0.345} & \textbf{0.448} & \textbf{0.534} \\
\midrule
ConvE & 0.290 & 0.207 & 0.316 & 0.456 & 0.281 & 0.200 & 0.309 & 0.442 \\
+SWA & 0.279 & 0.197 & 0.305 & 0.441 & 0.217 & 0.141 & 0.236 & 0.365 \\
+ASWA & \textbf{0.299} & \textbf{0.215} & \textbf{0.326} & \textbf{0.465} & \textbf{0.294} & \textbf{0.213} & \textbf{0.324} & \textbf{0.453} \\
\midrule
DistMult & 0.253 & 0.176 & 0.277 & 0.407 & 0.304 & 0.229 & 0.338 & 0.449 \\
+SWA & 0.264 & 0.188 & 0.285 & 0.419 & 0.268 & 0.187 & 0.302 & 0.426 \\
+ASWA & \textbf{0.285} & \textbf{0.206} & \textbf{0.307} & \textbf{0.445} & \textbf{0.398} & \textbf{0.331} & \textbf{0.432} & \textbf{0.526} \\
\midrule
TuckER & 0.321 & 0.234 & 0.350 & 0.494 & 0.196 & 0.126 & 0.209 & 0.333 \\
+SWA & 0.270 & 0.193 & 0.293 & 0.421 & 0.139 & 0.081 & 0.143 & 0.249 \\
+ASWA & \textbf{0.337} & \textbf{0.250} & \textbf{0.368} & \textbf{0.510} & \textbf{0.203} & \textbf{0.132} & \textbf{0.219} & \textbf{0.343} \\

\bottomrule
\end{tabular}
\end{table}